# Improving Disease Comorbidity Prediction Based on Human Interactome with Biologically Supervised Graph Embedding

Xihan Qin
Department of Computer and Information Sciences
*University of Delaware*
Newark, USA
xihan@udel.edu

Li Liao
Department of Computer and Information Sciences
*University of Delaware*
Newark, USA
liliao@udel.edu

*Abstract*—Comorbidity carries significant implications for disease understanding and management. The genetic causes for comorbidity often trace back to mutations occurred either in the same gene associated with two diseases or in different genes associated with different diseases respectively but coming into connection via protein-protein interactions. Therefore, human interactome has been used in more sophisticated study of disease comorbidity. Human interactome, as a large incomplete graph, presents its own challenges to extracting useful features for comorbidity prediction. In this work, we introduce a novel approach named Biologically Supervised Graph Embedding (BSE) to allow for selecting most relevant features to enhance the prediction accuracy of comorbid disease pairs. Our investigation into BSE's impact on both centered and uncentered embedding methods showcases its consistent superiority over the state-of-the-art techniques and its adeptness in selecting dimensions enriched with vital biological insights, thereby improving prediction performance significantly, up to 50% when measured by ROC for some variations. Further analysis indicates that BSE consistently and substantially improves the ratio of disease associations to gene connectivity, affirming its potential in uncovering latent biological factors affecting comorbidity. The statistically significant enhancements across diverse metrics underscore BSE's potential to introduce novel avenues for precise disease comorbidity predictions and other potential applications. The GitHub repository containing the source code can be accessed at the following link: https://github.com/xihan-qin/Biologically-Supervised-Graph-Embedding.

## I. INTRODUCTION

Comorbidity, the concurrent occurrence of multiple medical conditions within an individual [1], plays a pivotal role in disease management, treatment, and prognosis [2]. Understanding comorbidity patterns is essential for unraveling complex disease relationships and identifying shared molecular mechanisms [3]. In recent years, the study of disease comorbidities has gained considerable attention, leading to the development of network-based strategies for predicting and analyzing such relationships [2], [4], [5], [6], [7].

Beyond the simplistic approaches of finding common genes associated with different diseases as cause of comorbidity, the concept of a disease module was introduced by Menche et al. in 2015 [8] to examine whether and how protein-protein interaction may play a role in comorbidity by connecting genes associated with different diseases that do not share any common genes and therefore their comorbidity would be either missed in prediction or, even when identified clinically, lack of any explanation. Representing human interactome – collection of all protein-protein interactions in a human cell -- as a large graph with nodes for genes (or their protein product) and edges for interaction between proteins, a disease module is essentially a subgraph consisting of genes known to be associated with a disease. The authors in [16] introduced a method called disease module separation ($S_{AB}$), which measures the distance between two disease modules as shortest distance between a pair of nodes, with each node from the opposite disease module. Analyzing the data from a curated human interactome, the authors demonstrated that disease module separation exhibited a clear negative correlation with comorbidity, namely, the closer or shorter distance between the two modules are in the graph, the more likely the two diseases are comorbid. This pioneering work offered insights into exploring the incomplete human interactome for disease-disease relationships and beyond.

Interactome as a large and incomplete graph presents its own challenges to extracting useful features for more accurate comorbidity prediction. Graph embedding is a potent technique applied across domains like protein-protein interaction networks [9], social networks [10], and recommender systems [11], effectively captures significant graph connections in a low-dimensional Euclidean space [12]. A commonly used embedding type transforms the whole graph into vectors, with each vector representing a node [13], including methods like spectral embedding [14], node2vec [15], and graph convolutional neural network [16]. In Akram and Liao [17], an embedding method was proposed to incorporate graph structural information into comorbidity prediction. Utilizing the same human interactome data, their approach outperformed disease module separation, achieving a substantial 1.24-1.65-fold enhancement. Among their tests, dimension reduction method Isomap [18] with a selected dimension of 20, and the SVM RBF classifier provided the best performance through 10-fold cross-validation for predicting disease pair comorbidity.

In this work, we introduce a novel technique, Biologically Supervised Graph Embedding (BSE), designed to automatically select biologically pertinent embeddings for comorbid disease pair prediction. Through extensive analysis, we assess BSE's impact on enhancing prediction accuracy across both centered and uncentered embedding methods. Our results reveal that, using the same centered embedding (Isomap) as a starting point, our BSE consistently outperforms Akram and Liao's geodesic embedding method across all valid metrics. We elucidate how BSE augments predictive capabilities by selecting dimensions infused with vital "biological meaning". Moreover, our findings illustrate that BSE amplifies disease associations and gene connections with distinctive preferences for centered or uncentered embedding methods, thereby refining prediction performance.

Importantly, for all embedding techniques, the ratio of disease associations to gene connections improves significantly—up to 4 times—when using BSE compared to the singular value rank method. Through showcasing consistent enhancements across diverse performance metrics, we underscore how BSE's capacity to unveil latent "biological meanings" ushers in new possibilities for precise disease pair predictions and other potential applications.

## II. Related Work

Isomap used by Akarm and Liao [17] estimates the geodesic distance in a lower-dimensional space from a computed shortest distance matrix, D, among a selected number of neighbors or within a set radius of neighborhood in the graph. Then, multidimensional scaling (MDS) [19] is performed to obtain the final embedding. MDS uses D to estimate the centered Gram Matrix, $\bar{G}$, using the formula below.

$$\bar{G} = -\frac{1}{2} H D^2 H \quad (1)$$

Here, H is the centering matrix:

$$H = I - \frac{1}{n} J \quad (2)$$

With n being the number of nodes in the connected graph (the largest connected component of the human interactome), I the identity matrix of size n x n, and J the matrix of all 1s with size n x n.

The spectral decomposition of $\bar{G}$ generates the centered embedding matrix Z.

$$\bar{G} = U \Lambda U^T \quad (3)$$

$$Z = U_d \Lambda_d^{-\frac{1}{2}}, \ d < n \quad (4)$$

In the above equations, $U$ is a n x n matrix, consisted of orthonormal eigenvectors $u_1, u_2, \ldots, u_n$, and $\Lambda$ is a n x n matrix diagonal matrix with eigenvalues $\lambda_1, \lambda_2, \ldots, \lambda_n$ on the diagonal. The value of d is set by the user and represents the top d dimensions ranked by eigenvalues.

The geometric embedding [17] provides a novel way to measure the distance between the two disease modules A and B. Both $S_{AB}$ and geometric embedding aim to estimate the distance using shortest distance between nodes. However, $S_{AB}$ focuses on the shortest distance between two nodes from two different subgraphs, while geometric embedding estimates geodesic distance based on the shortest distances from all node pairs in the disease modules. The significant improvement observed with geometric embedding indicates that considering the shortest distances from all node pairs between disease modules, rather than just a single pair of nodes, is beneficial for describing disease distance in the context of comorbidity. This suggests that more hidden biological information is captured through geometric embedding by considering more nodes than using $S_{AB}$. However, given the inevitably large presence of noise in graph data, considering all node pairs between disease modules with selected dimensions based on eigenvalues may not be the most effective method for achieving the optimal biologically meaningful embedding. In this work, we set to look for improvement in identifying a more biologically meaningful approach to embedding the graph for disease pair comorbidity tasks.

## III. Materials

The human interactome data, comorbid disease pairs, and the regarding relative risk (RR) dataset were obtained from Menche et al. [8], which were also the datasets used by Akarm and Liao [17]. The human interactome comprises 13,460 gene IDs (graph nodes) that code for proteins. The largest connected subgraph contains 13,329 gene IDs, while other subgraphs consist of only a few nodes each. The curated protein-protein interaction types encompass regulatory, binary, literature, metabolic, complexes, kinase and signaling.

A total of 10,743 pairs of diseases with clinically reported RR scores were used. To ensure a fair comparison, we evaluated the datasets using two different thresholds: RR = 0 (dataset RR0) and RR = 1 (dataset RR1), following the approach employed by Akarm and Liao in their experiments [17]. In RR0 dataset, disease pairs with RR > 0 were marked as positive (i.e., comorbid, "1"), and disease pairs with RR ≤ 0 were marked as negative (i.e., non-comorbid, "0"). Similarly, in RR1, disease pairs with RR > 1 were marked as positive (i.e., comorbid), and those with RR ≤ 1 were marked as negative (i.e., non-comorbid). In dataset RR0, 82.6% of the pairs are positive, whereas in dataset RR1, 58.4% are positive.

## IV. Methods

Given the inherent incompleteness and noise in the human interactome, coupled with its high dimensionality, directly formulating a mathematical function to extract key "biological meanings" for the tasks is extremely challenging. Our proposed Biologically Supervised Embedding (BSE) method aims to leverage known knowledge which, in our case, are the human interactome and disease comorbidity labels.

Specifically, in contrast to an unsupervised embedding which uses eigenvectors ranked by their eigenvalues, we use a training dataset to select eigenvectors that contribute most in discriminating the comorbid disease pairs from non-comorbid disease pairs, and then use these selected eigenvectors for embedding to evaluate their performance on a reserved testing dataset.

Algorithm 1 details the steps of our BSE method. It takes a raw embedding matrix $Z_0$ as input, which can be obtained from any graph embedding method. The desired dimension number, d, is specified by the user and passed to the algorithm. Each dimension is selected through a cross-validation process, with the default set to 5 folds randomly chosen from our whole dataset. The feature vector for each disease pair is constructed following the approach described in [17], and the process is summarized briefly below.

Feature vector $F_i$ for disease i based on a given embedding matrix Z is represented as $[F_i^0 \ldots F_i^k \ldots F_i^{m-1}]$ with a length of m, which is the row size (length of columns) in Z. The value $F_i^k$ is calculated using the formula below:

$$F_i^k = \sum_{j \in S} Z_j^k \quad (5)$$

The value $F_i^k$ is calculated as the sum of the k-th column in the embedding matrix Z over the genes belonging to disease i (denoted as set S). For each disease pair with diseases *a* and

$b$, the feature vector becomes [$F_a^0$ … $F_a^k$ … $F_a^{m-1}$, $F_b^0$ … $F_b^k$ … $F_b^{m-1}$] with a length of 2m.

In BSE, the embedding matrix Z is constructed in a supervised learning manner iteratively, optimizing the prediction performance of a chosen task, in our case, detecting comorbid disease pairs. Therefore, unlike the conventional approach of choosing the top d columns ranked by eigenvalues to form the embedding matrix, we select columns by their contributions to maximizing the prediction performance (in our case, measured by AUC-ROC). If only embedded to one-dimension, we sequentially search through all n columns in the input embedding $Z_0$ to find which column form the best n x 1 embedding matrix Z, namely giving the highest performance. In cases where multiple dimensions yield the same optimal performance, we randomly select one of them. To increment to a two-dimension embedding, we would sequentially search the remaining n-1 columns to find one that can, when appended to the existing Z to form a new n x 2 embedding matrix, maximize the performance. Iteratively, we add more dimensions one at a time until n x d embedding matrix Z is constructed. This approach allows us to iteratively refine and select the most informative vectors, gradually revealing the key biological meanings relevant to the task.

**Algorithm 1**: Biologically Supervised Embedding

**Input:** M – a matrix, the collection of original vectors, d—desired number of vectors

**Output:** out_vec – the selected vectors

ran_id = random permutation of M's indices

auc_avg_ori = 0          # initialize the original average AUC score

# initialize the variable for selected vectors for output

out_vecs = empty column vector

for i = 0, 2, ……, d-1:

   auc_delta = []   # initialize a list to store the difference of AUC score

   for j = 0, 1, ……, len(rand_id)-1:

     ran_vec = M[rand_id[j]]         # randomly select one vector

     # construct candidate vectors

     cand_vecs = column_wise_append(out_vecs, rand_vec)

     # construct feature vectors for the task

     feature_vecs = construct_feature_vectors(cand_vecs)

     # calculate the average score

     avg_score = classifer_cross_validation (feature_vecs, labels)

     # store the score difference

     auc_delta.append((avg_auc - avg_auc_ori))

   **End for**

   # if there are more than one max number, randomly pick one

   max_idx = the index for max(auc_delta)

   # collect vectors for output

   out_vecs = column_wise_append(out_vecs, M[: , max_idx])

**End for**

**Return** out_vecs

Table I describes the 6 different variations we designed to test our BSE algorithm (Algorithm 1) for comorbid disease pair prediction tasks. These variations include the Biologically Supervised Embedding (E1), Ranked Embedding (E2), Biologically Supervised Vectors (E3), Ranked Vectors (E4), Biologically Supervised Geometric Embedding (E5), and Geometric Embedding (E6) methods. The main objective of these experiments is to assess how our algorithm can enhance both the centered and uncentered embedding techniques in comorbid disease pair prediction.

TABLE I.    DESIGNED EXPERIMENT STEPS TO TEST BSE

| E1. Biologically Supervised Embedding Approach | E3. Biologically Supervised Vectors Approach | E5. Biologically Supervised Geometric Embedding Approach |
|---|---|---|
| 1. Obtain the largest connected graph, G, from the human interactome. <br> 2. Generate a shortest distance matrix, D, from G <br> 3. Perform SVD on D <br> 4. Generate a raw embedding matrix $Z_0 = U_{100}\Sigma_{100}$ <br> 5. Run Algorithm 1 using $Z_0$, and d = 20 to generate the final embedding matrix Z. | 1. Same as E1. <br> 2. Same as E1. <br> 3. Same as E1. <br> 4. Generate a raw embedding matrix $Z_0 = U_{100}$ <br> 5. Run Algorithm 1 using $Z_0$, and d = 20 to generate the final embedding matrix Z. | 1. Same as E1. <br> 2. Same as E1. <br> 3. Perform Isomap using D and set the dimension number to 100 to output a raw embedding matrix $Z_0$ with dimension 100. <br> 4. Run Algorithm 1 using $Z_0$, and d = 20 to generate the final embedding matrix Z. |
| **E2. Ranked Embedding Approach** | **E4. Ranked Vectors Approach** | **E6. Geometric Embedding Approach** |
| 1. Same as E1. <br> 2. Same as E1. <br> 3. Same as E1. <br> 4. Select top 20 singular vectors ranked by singular values. <br> 5. The final embedding matrix Z = $U_{20}\Sigma_{20}$. | 1. Same as E1. <br> 2. Same as E1. <br> 3. Same as E1. <br> 4. Same as E2. <br> 5. The final embedding matrix Z = $U_{20}$ | 1. Same as E1. <br> 2. Same as E1. <br> 3. Perform Isomap using D and set the dimension number to 20 to generate the final embedding matrix Z with dimension 20. |

In the centered embedding techniques (E5 and E6), we employ Isomap (equations 1-4) to generate the embedding. For the uncentered approach (E1, E2, E3, and E4), we explain the design as below.

To implement the uncentered approach, we start by performing singular value decomposition (SVD) on the shortest distance matrix D, which can be represented as

$$D = U\Sigma V^T \qquad (6)$$

Here, U denotes the left singular vectors, which are equivalent to the eigenvectors for $DD^T$, which are the same as $D^2$ in (1).

$$DD^T = U\Sigma^2 U^T \qquad (7)$$

Likewise, V represents the right singular vectors, which are equivalent to the eigenvectors for $D^TD$, which is the same as $D^2$ (1).

$$D^TD = V\Sigma^2 V^T \qquad (8)$$

Since D is a symmetric square real matrix for an undirected graph, $DD^T = D^TD = D^2$. Therefore, taking the spectral decomposition of $D^2$ is equivalent to performing SVD on D [20].

For embedding approach (E1 and E2), we apply SVD on D, and then take:

$$Z = U_d \Sigma_d, \; d < n \quad (9)$$

For vectors approach E3 and E4, we apply SVD on D and then take:

$$Z = U_d, \; d < n \quad (10)$$

We set the number of dimensions for the final embedding matrix to 20 for all variations. This choice provides satisfactory results for a fair comparison and does not significantly increase the computation time. For the same reason, we set the raw embedding matrix $Z_0$ to size n x 100 in variations E1, E3, and E5. When the input $Z_0$ has a dimension number exceeds 100, we select the initial 100 dimensions by eigenvalues to reconstruct $Z_0$. These numbers are subject to modification based on the desired results or computational efficiency.

To evaluate the performance of each variation, we then use Stratified 10-fold cross-validation on datasets RR0 and RR1. We employ the SVM classifier with the Radial Basis Function (rbf) kernel and set the parameter C to 3.5. Various metrics, including accuracy, precision, recall, f1, and AUC, are used to score the performance.

Through these experiments, we aim to identify the most effective approach for comorbid disease pair prediction and assess the extent to which our Biologically Supervised Embedding Algorithm improves both the centered and uncentered approaches.

## V. RESULTS AND DISCUSSION

The outcomes of our study, obtained through method E1, E2, E3, E4, E5, and E6, have been meticulously examined and compared to underscore the efficacy of our proposed Biologically Supervised Embedding (BSE) algorithm. These approaches are marked by the notations "emb_select" (or "emb_s" for short), "emb_rank" (or "emb_r"), "vect_select" (or "vect_s"), "vect_rank" (or "vect_r"), "iso_select" (or "iso_s"), and "iso_rank" (or "iso_r"). The methods marked with "select" or "s" utilized Biologically Supervised Embedding (BSE) for dimension selection, while those marked with "rank" or "r" used singular values or eigen values to rank and select dimensions.

To delve into the dimensions selected by BSE and to gain deeper insights, we extracted the first 5 selected dimensions from all 10-fold cross-validation folds and used their union to test and output the average score. The results for the three embedding methods are denoted as "iso_u," "embed_u," and "vect_u."

TABLE II. AVERAGE METRIC SCORES FOR RR0

| Metric | iso_r | iso_s | p_val | std | emb_r | emb_s | p_val | std | vect_r | vect_s | p_val | std |
|---|---|---|---|---|---|---|---|---|---|---|---|---|
| precision | 0.8644 | **0.9046** | 3.71E-09 | 0.0057 | 0.8260 | **0.9074** | 8.63E-12 | 0.0059 | 0.8352 | **0.9075** | 7.02E-11 | 0.0066 |
| recall | **0.9846** | 0.9717 | 5.92E-05 | 0.0058 | **1.0000** | 0.9742 | 2.25E-08 | 0.0045 | **0.9977** | 0.9743 | 6.49E-07 | 0.0061 |
| f1 | 0.9206 | **0.9369** | 1.05E-06 | 0.0045 | 0.9047 | **0.9396** | 2.15E-09 | 0.0047 | 0.9093 | **0.9397** | 1.81E-08 | 0.0052 |
| Accuracy | 0.8596 | **0.8919** | 3.69E-07 | 0.0078 | 0.8260 | **0.8966** | 5.59E-10 | 0.0081 | 0.8355 | **0.8967** | 5.35E-09 | 0.0091 |
| roc_auc | 0.6255 | **0.7424** | 4.43E-09 | 0.0170 | 0.5000 | **0.7511** | 5.13E-12 | 0.0172 | 0.5315 | **0.7512** | 5.59E-11 | 0.0196 |

TABLE III. AVERAGE METRIC SCORES FOR RR1

| Metric | iso_r | iso_s | p_val | std | emb_r | emb_s | p_val | std | vect_r | vect_s | p_val | std |
|---|---|---|---|---|---|---|---|---|---|---|---|---|
| precision | 0.6975 | **0.7262** | 1.07E-05 | 0.0104 | 0.5859 | **0.7335** | 4.08E-11 | 0.0127 | 0.6456 | **0.7370** | 4.18E-09 | 0.0132 |
| recall | **0.8250** | 0.8102 | 1.37E-02 | 0.0154 | **0.9900** | 0.8102 | 1.52E-10 | 0.0179 | **0.8649** | 0.8057 | 8.80E-07 | 0.0158 |
| f1 | 0.7558 | **0.7658** | 4.84E-03 | 0.0085 | 0.7361 | **0.7698** | 7.13E-06 | 0.0116 | 0.7393 | **0.7697** | 6.87E-06 | 0.0104 |
| Accuracy | 0.6889 | **0.7109** | 1.21E-04 | 0.0108 | 0.5859 | **0.7173** | 3.37E-10 | 0.0143 | 0.6440 | **0.7187** | 5.29E-08 | 0.0144 |
| roc_auc | 0.6616 | **0.6910** | 3.22E-05 | 0.0122 | 0.5048 | **0.6987** | 2.12E-11 | 0.0155 | 0.5997 | **0.7013** | 1.01E-08 | 0.0162 |

Table II and Table III serve as platforms for presenting the average metric scores of the 10-fold cross-validation for both the RR0 and RR1 datasets. Notably, the rank method exhibits higher recall scores than BSE method, but at the expense of considerably lower AUC scores. A striking example emerges from the "emb_r" method applied to the RR0 dataset, wherein predicted labels for both real "1"s and "0"s are "1"s. Despite achieving precision and accuracy scores of 0.826 and recall of 1, the AUC score languishes at 0.5, indicative of random guessing and poor classification performance. Consequently, we decided not to employ recall scores for evaluation purposes.

Since in BSE, we leverage AUC improvement as our benchmark for dimension selection, the application of BSE results in a remarkable enhancement of AUC by up to 50.23% (for emb_r vs emb_s in Table II). Across the other metrics including precision, f1, and accuracy, BSE also consistently yields improvements of up to 25.18%, 4.57%, and 22.43% respectively, coupled with statistically significant p values. Importantly, the uniform high scores achieved by all BSE methods across various embedding techniques indicate that BSE has the capability to select biologically meaningful dimensions, leading to enhanced predictive performance across different embedding methodologies.

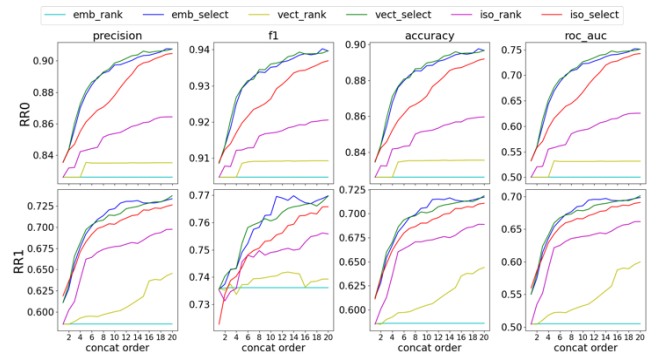

Fig. 1. Average metric scores along concatenated dimensions by BSE

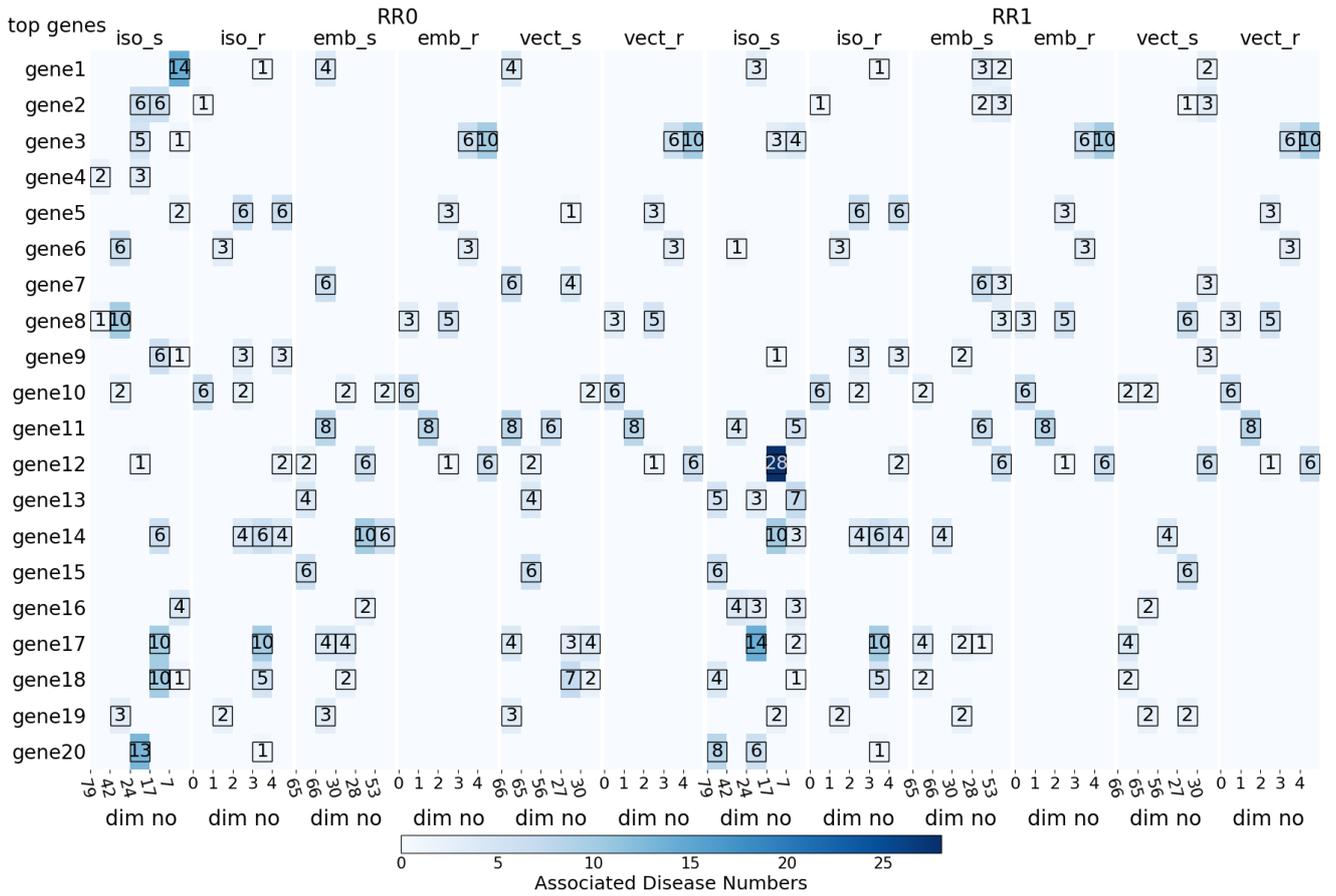

Fig. 2. Top 20 genes with associated disease numbers for first 5 dimensions.

Fig. 1 offers visual insights into the precision, f1, accuracy, and AUC average scores, exhibiting their progressive increments along the dimension concatenation order via BSE for both RR0 and RR1 datasets, compared to the rank method. The plots distinctly reveal substantial increments within the initial concatenations, often within the first 5, followed by gradual increases. Importantly, the dimensions selected by BSE markedly differ from the ranked dimensions, with limited overlap between the selected and ranked 20 dimensions. Moreover, the first 5 dimensions selected by BSE, consistently exhibit a conservative pattern across all 10 folds. Delving deeper, we explored the first 5 dimensions in the 10-fold cross-validation for all BSE methods. Employing the union of the first 5 dimensions for all 10 folds for each method, we performed tests and outputted the average score (Table IV). These scores closely mirror those obtained using the same embedding method with BSE. Intriguingly, the "vect_u" f1 score even surpasses that of "vect_s" for the RR1 dataset.

TABLE IV.  METRIC SCORES USING UNION DIMENSIONS

| Metric | RR0 | | | RR1 | | |
|---|---|---|---|---|---|---|
| | iso_u | emb_u | vect_u | iso_u | emb_u | vect_u |
| precision | 0.8863 | 0.8976 | 0.8901 | 0.7094 | 0.7377 | 0.7347 |
| recall | 0.9812 | 0.9772 | 0.9815 | 0.8264 | 0.8020 | 0.8095 |
| f1 | 0.9313 | 0.9357 | 0.9335 | 0.7634 | 0.7684 | **0.7702** |
| Accuracy | 0.8805 | 0.8891 | 0.8846 | 0.7012 | 0.7180 | 0.7182 |
| roc_auc | 0.6918 | 0.7240 | 0.7029 | 0.6761 | 0.7011 | 0.6999 |

To further explore for the biological insights, we picked the top 20 genes ranked by embedding values in the first 5 dimensions across all methods, we tallied the number of associated diseases for each gene (Fig. 2). The first 5 dimensions for methods employing BSE (marked as "s") were selected based on their frequency in the chosen 20 dimensions from all 10 folds. For comparative purposes with the singular value rank method, we designated dimensions by their respective singular values. For instance, in the case of the "iso_s" method, the first 5 selected dimensions were labeled as "79," "42," "24," "17," and "7," representing their original positions when ranked by singular values. The top 20 genes, ranked by their embedding values within each dimension, are displayed in corresponding columns. It is crucial to clarify that the "top genes" depicted in the plot vary across dimensions, methods, and datasets. For instance, "gene1" means the top 1 gene in each dimension. For the "iso_s" dimension "79" in the RR0 dataset, "gene1" is associated with the gene ID "387569," while for the "iso_r" dimension "0," "gene1" corresponds to the gene ID "126961." This pattern continues for other genes in each row. The color-coding signifies the number of associated diseases connected to each gene, with numbers in squares indicating gene-disease associations greater than 0. From the insights provided by Fig. 2, it is evident that numerous top genes lack known disease associations from the dataset. However, for the genes that do exhibit disease associations, the application of BSE significantly enhances the tally of disease associations for the centered approach ("iso_s"). On the other hand, for the uncentered approaches ("emb_s" and "vect_s"), this pattern displays a somewhat less distinct trend.

We further depict the gene association, another crucial facet of "biological meaning," by plotting the node degrees for the top 20 genes (Fig. 3). The spectrum of node degrees spans from 1 to 870, with varying colors representing these degrees—darker shades signify higher degree numbers. Interestingly, in the uncentered embedding approach, there is a discernibly lighter color pattern when utilizing BSE in comparison to the rank method. This phenomenon suggests that BSE aids uncentered embedding methods in identifying concealed and biologically significant connections, effectively mitigating connection "noise." However, this pattern is less conspicuous in the centered embedding method.

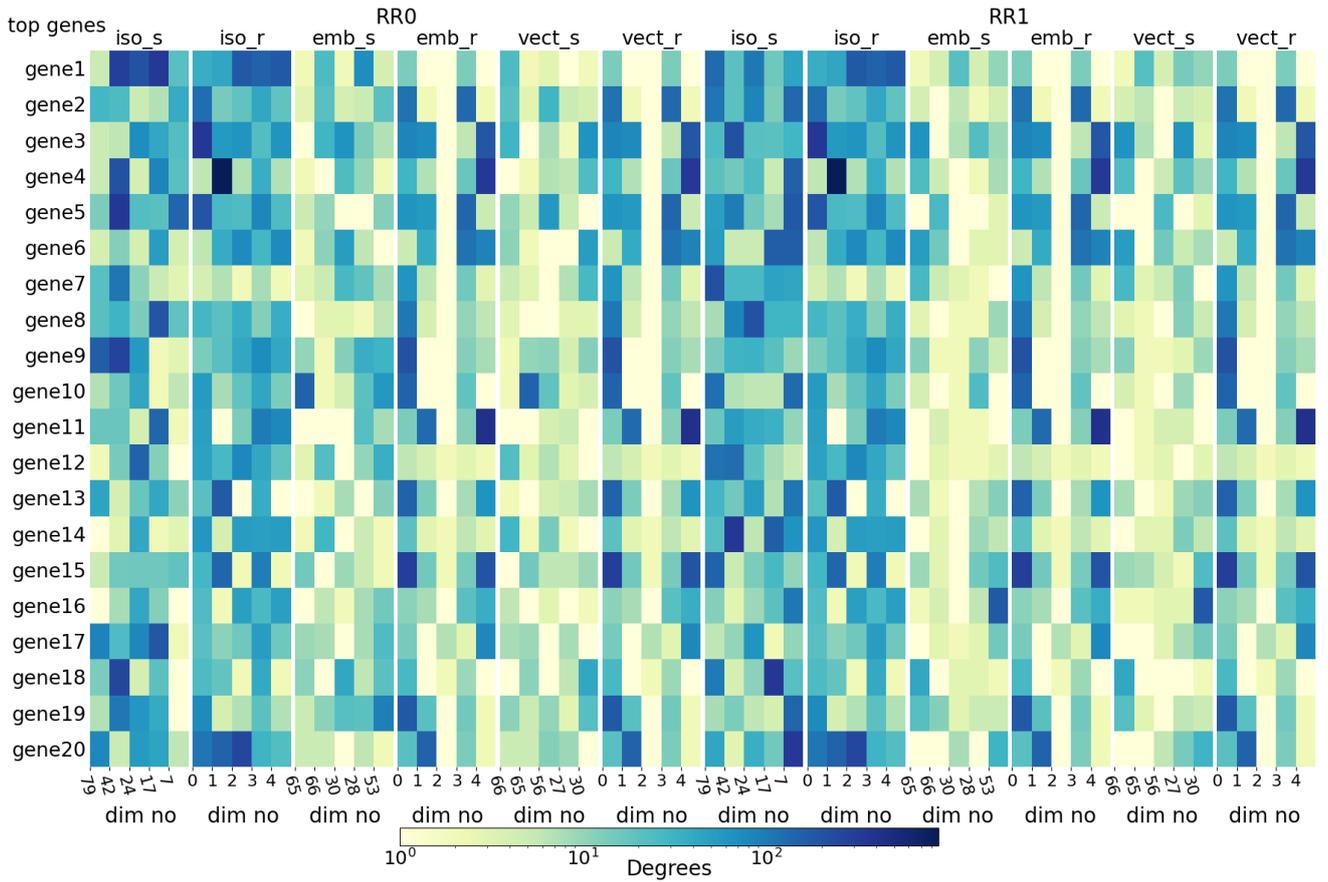

Fig. 3.    Top 20 genes and their degrees for the first 5 dimensions.

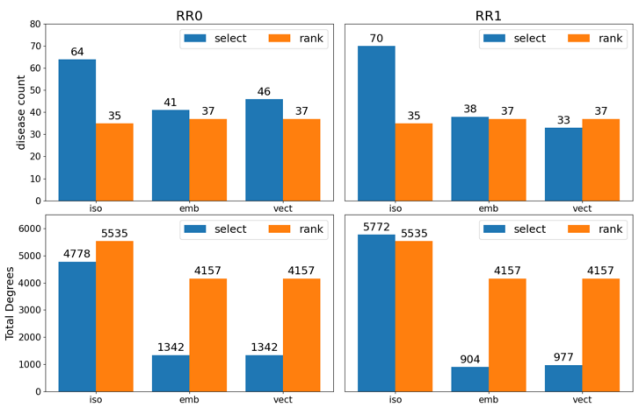

Fig. 4.    Disease count and degrees for top 20 genes and 5 dimensions

Furthermore, Fig. 4 presents comprehensive statistics for total disease count and total degrees associated with all top 20 genes within the initial 5 dimensions for all methods and both datasets. The cumulative disease count highlights an even more explicit trend: in centered approaches, BSE elevates the prominence of biological meaning—particularly disease associations—in the embedding matrix compared to the rank method. Conversely, for uncentered approaches, BSE accentuates the significance of gene association (total degrees) in the embedding matrix compared to the rank method. Notably, for both the "emb" and "vect" embedding methods, the total degrees experience a substantial reduction—3 to 4 times—when transitioning from the rank method to the BSE method. However, for the "iso" embedding method, the total disease count increases by up to 2 times for BSE compared to the rank method. These observations underscore the significance of both disease association and gene connectivity in this task.

More importantly, across diverse embedding methods, BSE demonstrates its ability to identify and incorporate essential biological meanings that play a pivotal role in achieving the task's objectives. This is manifested by a consistent enhancement of the metric R between disease association and gene connectivity, as defined by the formula below.

$$R = \frac{\sum_{i \in G, j \in D} s_i^j}{\sum_{i \in G, j \in D} d_i^j} \quad (11)$$

In this equation, the set D represents the chosen dimensions, wherein we selected the first 5 dimensions for all methods. Meanwhile, the set G represents the top genes within these dimensions, wherein we selected top 20 genes for evaluation. The variable $s_i^j$ denotes the associated disease number for gene i in dimension j, while $d_i^j$ signifies the degree of gene i in dimension j.

Table V provides an overview of the calculated ratios for all embedding approaches using BSE ("select") or the rank method. The results show a remarkable enhancement of approximately 2 to 4 times when utilizing BSE in comparison to the rank methods, indicating that when embedding the human interactome for disease comorbidity prediction, it is not disease association or gene connectivity alone, but rather their ratio (i.e., associating with more diseases with lower degree nodes in the network) that convey more biological relevance.

TABLE V. RATIO OF DISEASE ASSOCIATION AND GENE CONNECTIVITY FOR TOP 20 GENES IN FIRST 5 DIMENSIONS

|  | Iso | | Emb | | Vect | |
| --- | --- | --- | --- | --- | --- | --- |
|  | select | rank | select | rank | select | rank |
| RR0 | **0.0134** | 0.0063 | **0.0343** | 0.0063 | **0.0306** | 0.0089 |
| RR1 | **0.0121** | 0.0063 | **0.0376** | 0.0063 | **0.0399** | 0.0089 |

## VI. DISCUSSION

Our study aimed to enhance disease comorbidity prediction by leveraging biological information via an innovative Biologically Supervised Embedding (BSE) method. Through an extensive exploration of various embedding methods, both centered and uncentered, we conducted a thorough analysis of the impact of BSE on dimension selection and the subsequent performance improvement across multiple metrics.

The application of BSE, guided by AUC improvement, led to an impressive enhancement of up to 50.23% in AUC, highlighting its effectiveness in achieving a balance between sensitivity and specificity. Across other metrics, precision, f1, and accuracy metrics, BSE significantly contributes to the enhancement of prediction power, demonstrated consistent and substantial improvements with enhancements of up to 25.18%, 4.57%, and 22.43% respectively. These improvements were found to be statistically significant, reinforcing the robustness of our approach.

Our analysis of top genes in various embedding of human interactome has led to finding important biological insights. We observed that BSE not only enriched disease association counts, particularly in centered approaches, but also elucidated hidden biological connections in uncentered approaches. This ability to uncover and incorporate relevant biological meanings underscores the versatility and potential of BSE across diverse embedding methods. Moreover, our analysis reveals that, it is not the disease associations or gene connectivity alone, but rather their ratio that matters most while embedding the human interactome for disease comorbidity prediction.

Note that our algorithm in this work is a greed algorithm, namely, at each iteration, the best embedding is constructed by selecting one column from the remaining columns of the input raw matrix $Z_0$ to append to the best embedding from the previous iteration, and no backtrack is done to change the existing embedding. It is conceivable that further improvements can be achieved with more optimal embedding, of course at a higher computational cost.

## VII. CONCLUSION

In this work, we developed a novel Biologically Supervised Embedding approach to improve disease comorbidity prediction based on human interactome. The results from two benchmark datasets show the consistent and substantial enhancements across various performance metrics from the BSE approach as compared to conventional embedding techniques. Moreover, our study sheds light on the biological relevance of the ratio between disease associations and gene connectivity, not them individually, in affecting the predictive power of an embedding of human interactome. By unveiling hidden biological connections, our BSE approach opens avenues for more accurate and informed comorbid disease pair predictions. As BSE enables the extraction of biologically relevant patterns from complex data, its adoption could prove beneficial in various domains, such as protein interaction analysis, drug-target interaction prediction, and gene expression pattern recognition. While developed for analyzing biological networks, it is conceivable that this supervised embedding approach can also be adapted beyond bioinformatics for other applications where embedding is desirable analysis graph data.

ACKNOWLEDGMENT

The authors are grateful to Joerg Menche for making the comorbidity data available. The authors would also like to thank the anonymous reviewers for their invaluable comments.